%% file: main.tex
\documentclass[11pt]{article}
\usepackage[]{acl}

\input{math_commands.tex}

\input{Config/packages}
\input{Config/customized}

\title{\textit{From Informal to Formal} -- 
Incorporating and Evaluating LLMs on \\
Natural Language Requirements to Verifiable Formal Proofs
}

\author{
    {\bf Jialun Cao}$^{1}$,
    {\bf Yaojie Lu}$^{2}$,
    {\bf Meiziniu Li}$^{1}$,
    {\bf Haoyang Ma}$^{1}$,
    {\bf Haokun Li}$^{3}$, 
    {\bf Mengda He}$^{3}$, \\
    {\bf Cheng Wen}$^{3}$,
    {\bf Le Sun}$^{2}$,
    {\bf Hongyu Zhang}$^{4}$, 
    {\bf Shengchao Qin}$^{3,5}$,
    {\bf Shing-Chi Cheung}$^{1}$,
    {\bf Cong Tian}$^{5}$ \\
    $^1$ The Hong Kong University of Science and Technology \\
    $^2$ Institute of Software, Chinese Academy of Sciences \\
    $^3$ Guangzhou Institute of Technology, Xidian University  \\
    $^4$ Chongqing University 
    $^5$ ICTT and ISN Laboratory, Xidian University \\
    \texttt{jcaoap@cse.ust.hk}, \texttt{luyaojie@iscas.ac.cn} 
    \\\\
}

\begin{document}

\maketitle

\input{Tex/0-Abstract}
\input{Tex/1-Introduction}


\input{Tex/2-Problem}

\input{Tex/3-Dataset}
\input{Tex/4-Evaluation}

\input{Tex/7-Conclusion}
\input{Tex/8-Limitation}
\input{Tex/9-Acknowledgements}

\bibliography{reference}

\appendix
\input{Appendix/main-appendix}

\end{document}

%% file: math_commands.tex
\usepackage{amsmath,amsfonts,bm}

\def\eqref#1{equation~\ref{#1}}

\def\1{\bm{1}}

\DeclareMathAlphabet{\mathsfit}{\encodingdefault}{\sfdefault}{m}{sl}
\SetMathAlphabet{\mathsfit}{bold}{\encodingdefault}{\sfdefault}{bx}{n}



%% file: Config/packages.tex
\usepackage{times}
\usepackage{latexsym}

\usepackage[T1]{fontenc}

\usepackage[utf8]{inputenc}

\usepackage{microtype}

\usepackage{inconsolata}

\usepackage{graphicx}

\usepackage{hyperref}
\usepackage{url}

\usepackage{setspace}
\usepackage{bm}
\usepackage{xcolor}
\usepackage{microtype}
\usepackage{amsfonts}
\usepackage{xspace}
\usepackage{booktabs,hhline}

\usepackage{enumitem}
\usepackage{etoolbox}
\usepackage{array}
\usepackage{microtype}
\usepackage{multirow}
\usepackage{framed}
\usepackage{caption}
\usepackage{threeparttable}
\usepackage{fontawesome}
\usepackage[labelformat=simple]{subcaption}

\usepackage[ruled,vlined]{algorithm2e}

\usepackage{flushend}
\usepackage{balance}
\usepackage{graphicx}
\usepackage{amsthm}
\usepackage{amsmath}
\usepackage{amssymb}
\usepackage{bbding}
\usepackage{array}
\usepackage{multirow}
\usepackage{listings}
\usepackage{color}
\usepackage{tcolorbox}
\usepackage{xcolor,colortbl}
\usepackage{enumitem}
\usepackage{adjustbox}
\usepackage[normalem]{ulem}
\usepackage{pifont}
\usepackage{wasysym}

\usepackage{csquotes}

%% file: Config/customized.tex
\newcommand{\fma}{\textsc{fm-alpaca}\xspace}
\newcommand{\fmb}{\textsc{fm-bench}\xspace}

\setlist[itemize]{leftmargin=*}
\setlist[enumerate]{leftmargin=*}
\newlist{steps}{enumerate}{1}

\newcommand{\eg}{\emph{e.g., }}
\newcommand{\ie}{\emph{i.e., }}

%% file: Tex/0-Abstract.tex
\begin{abstract}
The research in AI-based formal mathematical reasoning has shown an unstoppable growth trend. These studies have excelled in mathematical competitions like IMO and have made significant progress. However, these studies intertwined multiple skills simultaneously—problem-solving, reasoning, and writing formal specifications—making it hard to precisely identify the LLMs' strengths and weaknesses in each task. This paper focuses on formal verification, an immediate application scenario of formal reasoning, and breaks it down into sub-tasks. 
We constructed {{18k high-quality instruction-response pairs}} across five mainstream formal specification languages (Coq, Lean4, Dafny, ACSL, and TLA+) in six tasks by distilling gpt-4o and evaluated against ten open-sourced LLMs, including recent popular DeepSeek-R1. 
We found that LLMs are good at writing proof segments when given either the code, or the detailed description of proof steps. Also, the fine-tuning brought about a nearly threefold improvement at most. 
Interestingly, we observed that fine-tuning with formal data also enhances mathematics, reasoning, and coding capabilities.
Fine-tuned models are released to facilitate subsequent studies at \url{https://huggingface.co/fm-universe}.

\end{abstract}

%% file: Tex/1-Introduction.tex
\section{Introduction}


\begin{displayquote}
{\footnotesize \ding{125}} \textit{The more we formalize, the more of our implicit knowledge becomes explicit.} {\footnotesize \ding{126}}

\hspace*{0pt}\hfill ---  Terence Tao~\cite{aimath-tao}
\end{displayquote}


\input{FigureTex/application}

As AI-based formal mathematical reasoning reached an inflection point~\cite{yang2024formal}, significant attention and progress in this field have been observed. AlphaProof~\cite{alphaproof2024ai} achieved silver medal level in the International Mathematical Olympiad (IMO), AlphaGeometry~\cite{alphageometry24} specialized in proving Euclidean geometry theorems. As reported, the number of publications in this field nearly doubled in 2023, indicating an unstoppable growth trend~\cite{li2024survey}. As Fields Medalist Terence Tao imagined, ``\textit{In the future, instead of typing up our proofs, we would explain them to some GPT}''~\cite{aimath-tao}.

However, most current benchmarks cannot precisely reflect the capability to convert informal proofs or requirements in natural language into formal proofs. 
Most of these benchmarks take mathematical problems~\cite{alphaproof2024ai,alphageometry24,welleck2021naturalproofsmathematicaltheoremproving} or theorems to be solved~\cite{yang2023leandojo,welleck2022naturalprover,coqgym} as input, and informal or formal proofs (or parts of proofs) as output. 
However, these end-to-end benchmarks assess multiple capabilities (\eg problem-solving, mathematical reasoning, formal specification writing) {{in an intertwined manner}},
{{making it difficult to isolate and observe LLMs' true capabilities in writing formal proofs or models for verification.}}

Therefore, we \textbf{\textit{break down the process}} from informal requirements to formal verifiable proof, as shown in Figure~\ref{fig:motivation}. Inspired by the code generation (shown in blue) which translates a description of implementation into executable code~\cite{humaneval,mbpp2021}, the formal reasoning process (shown in green) can be seen as translating an \textbf{\textit{informal requirement}} into \textbf{\textit{a verifiable formal proof}} or checkable formal model~\footnote{For ease of expression, we generally refer to {\textit{verifiable formal proofs}} and \textit{checkable models} as ``proofs'' for the sake of presentation simplification.\label{ft:term}}. Particularly, we decompose this process and formulate six tasks (Figure~\ref{fig:task}). By doing so, the intertwined capabilities can be separated and individually assessed, providing a clearer understanding of LLMs' strengths and weaknesses in each task.

\textbf{{Scope and Targets}} -- We focus on \textbf{\textit{formal verification}}~\cite{verifiedToolchain,sel4,leroy2016compcert,hawblitzel2014ironclad} because it is an immediate application scenario of formal mathematical reasoning and the correctness of the output can be verified mechanically. In this paper, we mainly explore four research questions (RQs):

\textbf{\textit{RQ1. How well do LLMs perform in various formal verification tasks?}} After decomposing the formal verification task into subtasks, we explore LLMs' initial performance in these tasks with zero-shot and few-shot, investigating the strengths and weaknesses that vary between LLMs and tasks.

\textit{\textbf{RQ2. Do LLMs show variability in their capability across different formal specification languages?}} 
When mathematicians and proof engineers consider using LLMs to assist in formal verification, they often face uncertainty about which formal specification language is best supported by LLMs. This RQ is designed to provide hints on it. 

\textbf{\textit{RQ3. Can fine-tuning improve LLMs' performance in formal verification}}? Although recent efforts have been made to fine-tune models~\cite{leanexpert24,yang2023leandojo}, these LLMs are typically fine-tuned with single formal languages instead of multi-lingual (e.g., combining Coq, Lean, etc.)~\cite{yang2024formal}. 
Therefore, we instruction fine-tuned~\cite{wei2021finetuned,sanh2021multitask} three base LLMs to see whether our constructed fine-tuning dataset \fma could improve their capability in formal verification tasks.

\textbf{\textit{RQ4. 
Can fine-tuning with formal verification data benefit 
other related tasks (mathematics, reasoning, code)?}} As recent works have shown LLMs' potential transferability of skills~\cite{tihanyi2023new} 
we thus extend our study to see if models fine-tuned on formal data could show enhanced capabilities in mathematics, reasoning, and coding.

To facilitate the study, we constructed \textbf{\textit{18k high-quality instruction-response pairs}} across five formal specification languages (\ie Coq, Lean4, Dafny, ACSL, and TLA+) in six formal-verification-related tasks by distilling gpt-4o inspired by prior work~\cite{leanexpert24,ultrachat23,selfinstruct22}, then split them into 14k instruction fine-tuning data (\fma) and 4k benchmarking data (\fmb). 
In particular, we provide \textbf{\textit{executable contexts}} for all these formal specifications and automated \textbf{\textit{validation scripts}} to validate the correctness of the generated formal proofs inspired from the prior work's artifact preparation~\cite{swe-bench}. Finally, we release the fine-tuned LLMs based on three base models at \url{https://huggingface.co/fm-universe}.

Interestingly, there has been recent discussion on the topic of domain transfer~\cite{yang2024formal}, particularly the transfer of knowledge {{from other domains such as coding and reasoning to formal domains}} in order to increase LLMs' reliability~\cite{spiess2024calibration}, and the anticipated potential of AI in enhancing formal verification processes to support mathematical proofs~\cite{aimath-tao,tao2023formalizinglean4}. {{Our experimental results could potentially provide empirical support for these hypotheses or offer directions for further experimental inquiries.}}


    
    

The contribution of this paper includes:

    \ding{52} \textbf{\textit{Problem Formulation}}: 
    We decompose the formal verification process into six essential tasks. By doing so, the intertwined capabilities can be separated and individually assessed, providing a clearer understanding of LLMs' strengths and weaknesses in each task. 
    
    
    \ding{52} \textbf{\textit{Dataset and Benchmark}}: 
    We constructed {{18k high-quality instruction-response pairs}} across five mainstream formal specification languages (\ie Coq, Lean4, Dafny, ACSL, and TLA+) in six formal-verification-related tasks by distilling gpt-4o. They are split into a 14k+ fine-tuning dataset \fma and a 4k benchmark \fmb. 

    \ding{52} \textbf{\textit{Executable context and automated validation mechanism}}: We provide a Docker container equipped with necessary scripts to facilitate the evaluation of \fmb, significantly lowering the entry barrier for this scenario and making subsequent contributions easier.

    \ding{52} \textbf{\textit{Insight and Vision}}: We fine-tuned several models on \fma and observed promising benefits to not only the formal verification tasks, but also mathematics, reasoning, and coding. Our experimental results provide empirical support for the potential of LLMs' capability transfer and hope to shed some light on future research.
    
    

%% file: FigureTex/application.tex
\begin{figure*}[th]
    \centering
    \includegraphics[width=1.0\textwidth]{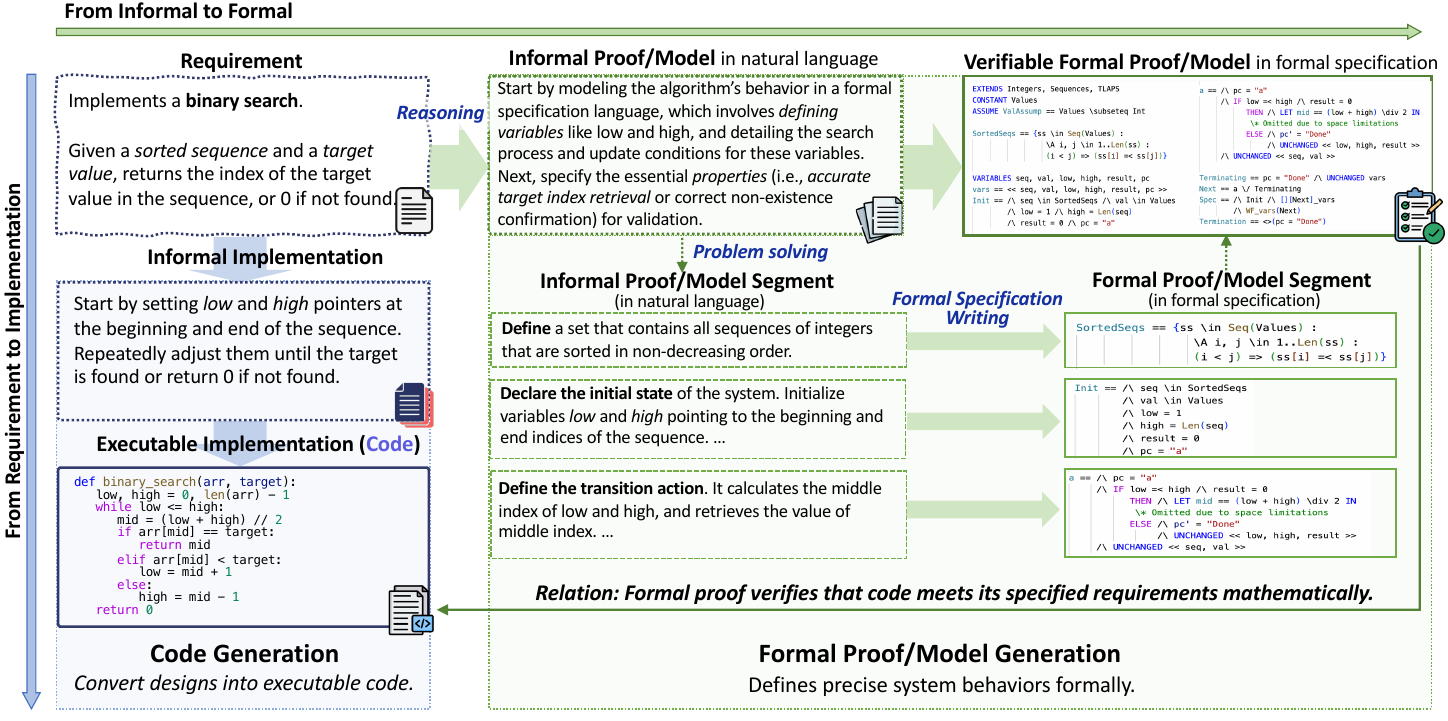}
    \setlength{\belowcaptionskip}{-10pt}
    \caption{The Illustration of Formal Proof Generation and Its Relation with Code Generation.}
    \label{fig:motivation}
\end{figure*}

%% file: Tex/2-Problem.tex
\section{Task Formulation}

Figure~\ref{fig:task} illustrates the six sub-tasks. We elaborate on them in detail as follows.

\textbf{\textit{Task 1. Requirement Analysis}} 
(abbrev. ReqAna). Requirement analysis~\cite{davis1990software-requirement,anton1996goal,grady2010system,jin2017environment} is a critical and long-standing research area in software engineering. It facilitates collecting, identifying, categorizing and modeling the users' needs and expectations using various techniques~\cite{taggart1977survey,req2uml09,javed2021imer,wang2021intelligent,jin2024evaluation,zhou2022assisting}. In this paper, the requirements are the descriptions in natural language (English)~\cite{jin2024evaluation} that details the requirements of the verification/modeling goal and an overall description of the proofs/models. The task is to analyze and break down the final goal into detailed steps described in natural language. The natural language used in this paper is English.

\textbf{\textit{Task 2. Full Proof Generation}} (abbrev. ProofGen).
This task formalizes a requirement in natural language into verifiable proofs or models written in formal specification languages, similar task formulation to existing works~\cite{fatwanto2012translating,zhou2022assisting,featuremodel13}. 

\textbf{\textit{Task 3. Proof Segment Generation}} (abbrev. SegGen). Unlike ProofGen, which requires generating complete proofs/models, SegGen provides more detailed descriptions in natural language and requires LLMs to write less. Given a text description articulating how to implement the proofs/modeling, the task outputs a segment written in the formal specification that serves as a component in the complete proof/model. This task formulation is similar to prior work~\cite{leanexpert24,wu2022autoformalization,jiang2022draft} and similar to the formulation of code generation~\cite{codealpaca,sun2024cloverclosedloopverifiablecode,welleck2022naturalprover,welleck2021naturalproofsmathematicaltheoremproving}. 

\textbf{\textit{Task 4. Proof Completion}} (abbrev. ProofComplete). Similar to code completion~\cite{raychev2014codecompletion,codecompletion-survey,svyatkovskiy2019pythia,dakhel2023github}, ProofComplete suggests the suffix of the given prefix, similar to prior work~\cite{song2024towards}. Note that in order to prevent LLMs from deviating from the original verification goal, we also provide the requirement in our evaluation, although it is not compulsory for this task formulation. 

\textbf{\textit{Task 5. Proof InFilling}} (abbrev. ProofInfill). Given a proof/model with a mask in the middle, the task requires LLMs to fill proper formal specifications so that the completed proofs/models can pass the verifier. This formulation is the same as code infilling~\cite{fried2022incoder}. Also, similar to ProofComplete, we provide the requirement in our evaluation during the infilling to prevent LLMs from deviating from the original verification goal.

\textbf{\textit{Task 6. Proof Generation from Code}} (abbrev. Code2Proof). In addition to generating formal specifications from natural languages, formal specifications can also be generated from code if the verification goal is the property of a given program. In this paper, we focus mainly on specifications in form of code annotations~\cite{baudin2021acsl,hatcliff2012behavioral}, expressing specifications (\eg pre-/post-condition, loop invariants) that help one to verify that (part of) a program satisfies certain properties. The task takes the code with properties to be verified as input and outputs the code with generated annotated formal specifications. 
Similar task formulation can be found in recent works~\cite{autospec24,ma2024specgen}.

\input{FigureTex/task}


%% file: FigureTex/task.tex
\begin{figure*}[th]
    \centering
    \includegraphics[width=1.0\textwidth]{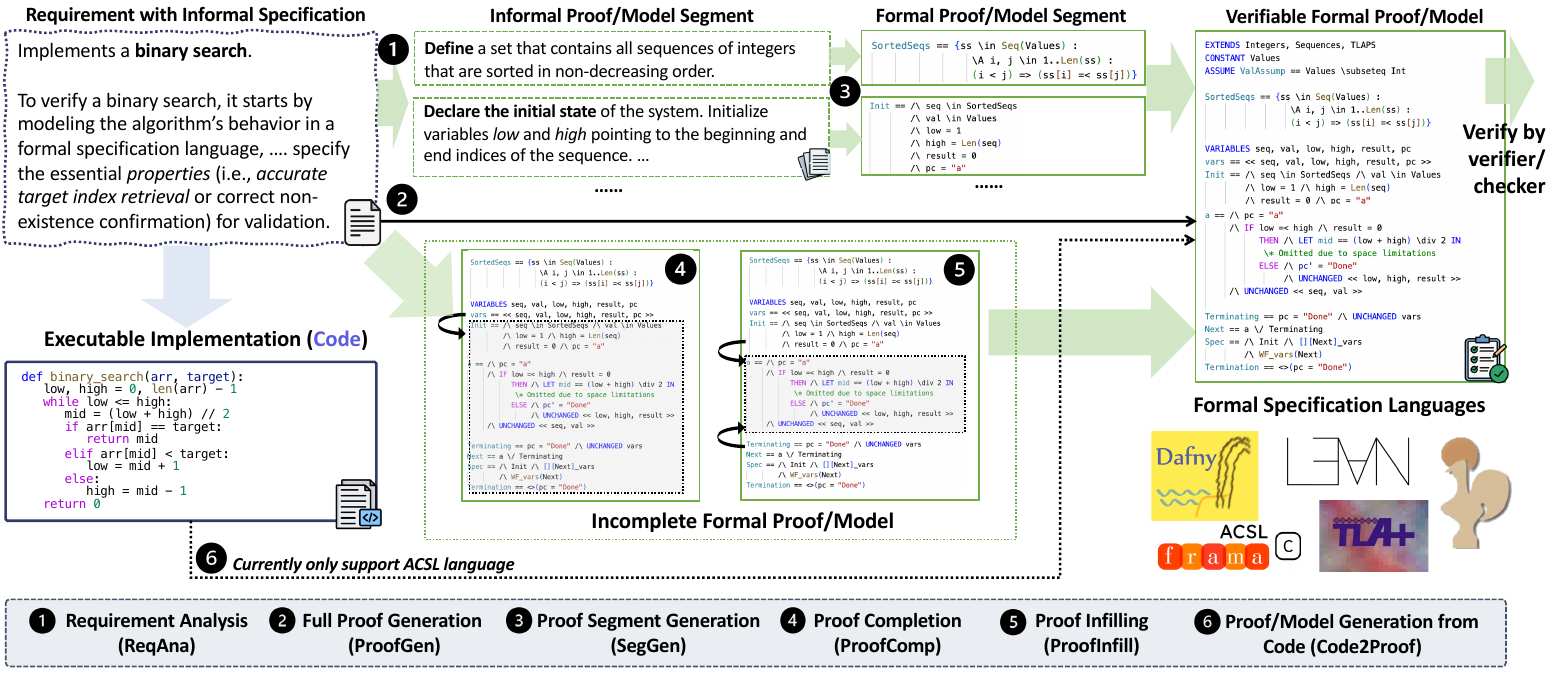}
    \setlength{\belowcaptionskip}{-10pt}
    \caption{Six tasks towards Informal to Formal Verification}
    \label{fig:task}
\end{figure*}

%% file: Tex/3-Dataset.tex
\section{Data Construction}\label{sec:construct}

\subsection{Formal Specification Language Selection}
In this study, we consider five formal specification languages that can be used for formal verification, including Coq~\cite{coq}, Dafny~\cite{leino2010dafny}, Lean4~\cite{lean4}, ACSL (ANSI/ISO C Specification)~\cite{baudin2021acsl} and TLA+~\cite{tla1999,tla2002}. 
We selected them in order to cover various \textbf{\textit{verification paradigms}} (\ie theorem proving and model checking) and \textbf{\textit{verification scenarios}} (\eg mathematical reasoning and program verification). 

First, for \textbf{\textit{interactive theorem provers}} which are suitable for developing rigorous mathematical proofs, we consider \textbf{Coq}~\cite{coq} and \textbf{Lean4}~\cite{lean4} because Coq has been extensively used in academia and research for proving mathematical theorems and in formal verification of software for a long history, while
Lean4 garnered considerable attention from the mathematical community~\cite{leanexpert24,tao2023formalizinglean4,avigad2020mathematics} recently. 
Second, for \textbf{\textit{programming languages with built-in specification}}, we consider Dafny~\cite{leino2010dafny} and ACSL~\cite{baudin2021acsl,frama-c} because they seamlessly integrate specifications (\eg pre-/post-conditions, loop invariants) within the code, ensuring the correctness through embedded assertions and conditions.
Lastly, for \textbf{\textit{model checking}}~\cite{jhala2009softwaremodelchecking,clarke1997model}, we consider \textbf{TLA+}~\cite{tla1999} since it is a representative math-based formal language for modeling algorithms and programs such as concurrent and distributed systems.

\subsection{Data Preparation}

\input{FigureTex/data-pre}

The workflow of data preparation for \fma and \fmb is illustrated in Figure~\ref{fig:data}. 
The workflow begins with the data collection, where formal proofs in the desired formal specification languages and related configurations and dependencies are gathered from open-source repositories in Github. 
Then, formal proofs are \textbf{{extracted}} from the collected repositories. Next, the proofs go through the data quality assurance check by execution, the proofs that cannot be verified successfully are filtered out. The remaining ones are \textbf{{split}} into segments (\eg definition of functions or conditions). 

Given the impracticality of manually writing descriptions for all the collected formal proofs,
we leveraged \textit{distill}ed GPT4~\cite{gpt4turbo} to generate high-quality informal proof descriptions via meticulous prompting. This alternative is well-established and frequently employed in prior literature~\cite{selfinstruct22,leanexpert24,ultrachat23,tuluv1}.
Specifically, for each formal specification language, we designated the model as an expert in that particular language, equipping it with comprehensive domain knowledge about the language's specifications, essential grammatical cues, and three-shot examples featuring proof segments in the formal language as inputs and natural language descriptions as outputs. This approach ensures that the collected descriptions are of high quality and well-organized. It's important to note that we did not generate descriptions for proof segments shorter than two lines (such as package imports or constant definitions) because their meaning is self-explained, with the exception of ACSL, whose proof segments are typically 1-2 lines.

After the descriptions for both full and segment proofs were prepared, we then prepared the data pairs for each task as shown in \textbf{Task-wise Data Pairing} in Figure~\ref{fig:data}. Note that for Task 4, \ie Proof Completion, to prepare the incomplete formal proof, we randomly choose a line number and delete the lines in the proof after the line. For Task 5, \ie Proof Infill, we randomly choose two line numbers and mask the lines between them. In case the remaining lines of proof cannot provide sufficient information for the proof generation, we also provide the informal proof for these two tasks. 

After pairing the instruction-response for different tasks, we manually designed five task instructions for each task and randomly assigned one for each paired data to increase instruction diversity and avoid overfitting to certain instructions~\cite{lu2022prompt,feng2023diverseprompt,sanh2022multitaskpromptedtrainingenables}.



\input{TableTex/num-stats}

\subsection{Data Statistics}
The specification-language-wise and task-wise statistics are shown in Table~\ref{tab:num-stats} and Table~\ref{tab:num-tasks}. 
In particular, Table~\ref{tab:num-stats} presents a detailed breakdown of the number of proofs and segments across five specification languages. 
Note that we split the prepared data for all the tasks and specification languages into an 8:2 ratio, \ie 80\% for fine-tuning, named \fma, 20\% for benchmarking, named \fmb, and show the separate statistics. 
In particular, there are 4k+ verifiable proofs in total, with 249 $\sim$ 2k proofs for each language. These proofs were split into 18k+ proof segments, with an average of 3.6k segments for each language. The reason why the ratio of segments in \fma and \fmb is slightly less than 8:2 is that the train-test split was applied to proofs, while the number of split segments in each proof varies.



\input{TableTex/num-task}

Table~\ref{tab:num-tasks} shows the task-wise statistics. There are 18k instructions across six tasks, with an average of 3k instructions for each task. After splitting the train-test set, \fma contains 14k, and \fmb has nearly 4k instructions. It is clear that the number of instructions for the task Segment Proof Generation (SegGen) is far more than that for Requirement Analysis (ReqAna) and Full Proof Generation (ProofGen) because one full proof can be split into numerous pieces of proof segments, and one proof can contribute to only one instruction for ReqAna and ProofGen. Note that the number of ReqAna (627) and ProofGen (700) is unequal because we filtered out the instructions with more than 2048 tokens considering the context limits.






\subsection{Validation Mechanism}
For the tasks whose outputs are written in formal specification languages, we verify the full proofs against the corresponding verifiers, \ie Coq, Dafny, Lean4 use their own proving environment; formal specification written in TLA+ can be checked by TLC~\cite{tlc}; C programs with ACSL specifications can be checked by Frama-C~\cite{frama-c,carvalho2014formal}. Also, for the proof segments that cannot be verified independently, for each extracted segment, we prepared a proof template with a placeholder during data preparation. Whenever a generated segment is to be verified, we replace the placeholder in the template with the segment and verify the completed formal proof. 

For ACSL specification verification, we report the results under two modes. First, \textbf{Basic checking}. ACSL uses the `frama-c' tool for syntax checking (\ie, whether the syntax is correct) and other basic ACSL checking. Second, \textbf{Verification} which involves semantics checking (\ie program behaviors are consistent with the annotated ACSL specifications) and verification towards the property under verification. `ACSL-WP' uses `frama-c' with the {WP} plugin for formal verification. This mode requires not only the basic checking (\ie, syntactically correct, semantically consistent with the program), but also that the properties under verification can be proved.

Note that the invalid outputs (\eg empty response or responses without formal specifications) are considered incorrect directly and will not go through the verification.

For the task whose outputs are written in natural language (\ie ReqAna), we calculate the Bleu score~\cite{bleu-score2002} between the descriptions in \fmb with the predicted outputs. 

\input{TableTex/pass1}

%% file: FigureTex/data-pre.tex
\begin{figure}[t!]
    \centering
    \includegraphics[width=1.0\linewidth]{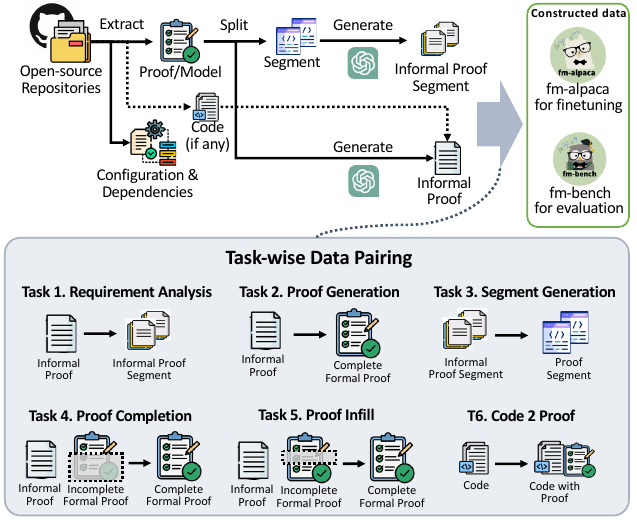}
    \setlength{\abovecaptionskip}{-0pt}
    \setlength{\belowcaptionskip}{-10pt}
    \caption{The Illustration of Data Preparation.}
    \label{fig:data}
\end{figure}

%% file: TableTex/num-stats.tex
\begin{table}[t!]
    \centering
    \resizebox{1.0\linewidth}{!}{

\begin{tabular}{l|r|rr||r|rr}
\toprule
\multicolumn{1}{l|}{\multirow{2}{*}{Spec}} & \multicolumn{3}{c||}{Num of Proofs} & \multicolumn{3}{c}{Num of Segments} \\
\multicolumn{1}{c|}{} & Total & \fma & \fmb & Total & \fma & \fmb \\
\midrule
Coq & 2,126 & 1,683 & 443 & 14,939 & 11,638 & 3,301 \\
Lean4 & 1,163 & 919 & 244 & 1,578 & 1,261 & 317 \\
ACSL & 544 & 426 & 118 & 765 & 598 & 167 \\
Dafny & 249 & 206 & 43 & 417 & 348 & 69 \\
TLA+ & 256 & 199 & 57 & 594 & 476 & 118 \\
\midrule
\textbf{Total} & 4,338 & 3,433 & 905 & 18,293 & 14,321 & 3,972 \\
\textbf{Average} & 868 & 687 & 181 & 3,659 & 2,864 & 794 \\
\bottomrule
\end{tabular}
    }
    \caption{Formal-Specification-Language-wise Statistics of Formal Verification Data}
    \label{tab:num-stats}
\end{table}

%% file: TableTex/num-task.tex
\begin{table}[t!]
    \centering
    \resizebox{1.0\linewidth}{!}{

\begin{tabular}{llrrr}
\toprule
 & Task & Total & \fma & \fmb \\
 \midrule
1 & Requirement Analysis & 627 & 496 & 131 \\
2 & Full Proof Generation & 700 & 557 & 143 \\
3 & Segment Proof Generation & 14843 & 11597 & 3246 \\
4 & Proof Complete & 658 & 520 & 138 \\
5 & Proof Infill & 1439 & 1146 & 293 \\
6 & Code2Proof & 70 & 56 & 14 \\
\midrule
 & \textbf{Total} & 18337 & 14372 & 3965 \\
 & \textbf{Average} & 3056 & 2395 & 661 \\
\bottomrule

\end{tabular}
}
\caption{Task-wise Statistics of Formal Verification Data}
\label{tab:num-tasks}
\end{table}

%% file: TableTex/pass1.tex
\begin{table*}[ht]
    \centering
    \renewcommand\arraystretch{1.3}
    \resizebox{1\linewidth}{!}{
\begin{tabular}{rrrrrrrrrrrrrrrrrrrrrrrrrrr}
\toprule
\multicolumn{1}{c}{}                                & \multicolumn{1}{c}{}                                & \multicolumn{6}{c}{\textbf{ProofGen}}                                                                                                                                                                    & \multicolumn{6}{c}{\textbf{SegGen}}                                                                                                                                                              & \multicolumn{4}{c}{\textbf{ProofComplete}}                                                                                                         & \multicolumn{6}{c}{\textbf{ProofInfill}}                                                                                                                                                                         & \multicolumn{2}{c}{\textbf{Cd2Prf}}                                             & \multicolumn{1}{c}{}                      \\
\end{tabular}

    }
    \caption{\textbf{RQ1-3: Pass@1 Accuracy of LLMs' Performance Across Formal Verification Task and Formal Specification Languages} with (w/) and without (w/o) fine-tuning. The \colorbox[HTML]{C5E7E5}{greener}, the better. }
    \label{tab:pass1}
\end{table*}

%% file: Tex/4-Evaluation.tex
\section{Experiments}

\subsection{Experiment Setup}

\noindent \textbf{\textit{Studied LLMs}}. We selected ten LLMs as baselines without fine-tuning, including llama3.1-instruct-8B/70B~\citep{meta-llama-3}, qwen2.5-instruct-7B/72B~\citep{qwen2.5-report}, qwen2.5-coder-instruct-7B/-32B~\cite{qwen2.5-coder}, starcoder-instruct-15B~\cite{lozhkov2024starcoder}, deepseek-coder-instruct-7B-v1.5, deepseek-coder-instruct-33B~\citep{deepseek-coder}, and deepseek-R1~\cite{dsr1}. Note that we avoid evaluating the GPT-series LLMs by OpenAI because the descriptions in \fmb were generated by GPT-4o, making the evaluation fairer.

\textbf{\textit{Fine-tuning}}. 
Instruction fine-tuning~\cite{wei2021finetuned,sanh2021multitask,ultrachat23,tulu2} aims to improve a model’s ability to effectively respond to human instructions and has shown strong experimental potential in model enhancement. 
 We select llama3.1-8B-instruct~\cite{meta-llama-3}, qwen2.5-7B-instruct~\cite{qwen2.5-report}, and deepseek-coder-7B-instruct-v1.5
 ~\cite{deepseek-coder} as base models for fine-tuning.
These three models are chosen due to their promising performance on tasks such as coding, mathematics, and reasoning, and fine-tuning models in their scale is relatively affordable compared with fine-tuning larger scale models.
We fine-tuned the three aforementioned models over three epochs using a learning rate 2e-5, a warm-up ratio of 0.04, a batch size of 512, and a cosine learning rate scheduler.

\input{TableTex/reqsplitter}

\textbf{\textit{Baseline Fine-tuning Datasets}}: To distinguish whether the capability improvement is simply because more instruction tuning is applied, we also include two commonly used fine-tuning datasets for comparison.
We select UltraChat~\cite{ultrachat23} and Tulu-V3~\cite{tulu3} as baseline fine-tuning datasets, and use llama3.1–8B-base~\cite{meta-llama-3} as the base model due to their popularity.
In particular, UltraChat is a large-scale dataset of instructional conversations that contains 1.5 million high-quality multi-turn dialogues and covers a wide range of topics and instructions. Tulu-v3~\cite{tulu3} embraces new data that is either carefully manu   ally curated for quality or generated from GPT models. 
It is an enhancement of its previous versions~\cite{tulu2,tuluv1}, focusing more on core skills of knowledge recall, reasoning, mathematics, coding, instruction following, general chat, and safety. 

\textbf{\textit{Benchmarks for Related Capabilities (RQ4).}}
To comprehensively evaluate the model's capabilities, we tested the fine-tuned models on a series of benchmarks: Math~\citep{hendrycksmath2021} and GSM-8K~\citep{cobbe2021gsm8k} for mathematical reasoning, BBH~\citep{suzgun2022challenging} for general reasoning,  HumanEval~\citep{humaneval} and MBPP~\citep{austin2021program} for coding.

\textbf{\textit{Inference Strategies}}: We adopt different settings for different RQs. In particular, 
We use (1) the greedy sampling strategy to generate one single greedy sample with a temperature of 0.0 and calculate Pass@1, and (2) nucleus sampling~\cite{nuclear}, where five solution samples were randomly generated with a temperature of 0.2 for RQ1 and RQ2. We also consider different in-context learning strategies, including zero-shot and few-shot (we used 3-shot in the experiment). For RQ3, we use a zero-shot greedy search with a temperature of 0.0 and a few-shot nuclear search with a temperature of 0.2 for a fair comparison.

\textbf{\textit{Experiment Environment}}.
The fine-tuning experiment was conducted on 32 Nvidia A800-40G GPUs, while inference was on Nvidia A100-80G GPUs with vLLM \citep{kwon2023efficient}.

\subsection{RQ1. Basic Performance across Formal Specification Tasks}\label{sec:rq1}
To understand the current LLMs' performance in six tasks, we evaluate 10 LLMs against \fmb with model size ranges from 7B to 671B. The upper part of Table~\ref{tab:pass1} and the upper part of Table~\ref{tab:reqsplitter} show LLMs' basic performance without fine-tuning. 

\textbf{Task-wise:} LLMs perform the best in ProofComplete (18.31\%) and Code2Proof (16.79\%). In contrast, LLMs fall short in generating both the entire formal proof (7.70\%) and the proof segments (9.16\%). We analyzed the failures and found that syntax errors account for a large proportion, with 12.43\% failures caused by syntax errors (See Appendix~\ref{sec:syntax_error}).
The observation echoes the motivation of prior work~\cite{leanexpert24} and is reasonable due to the grammar difference between most formal specification languages and other programming languages like Python. Regarding requirement analysis, as shown in the upper part of Table~\ref{tab:reqsplitter}, the Bleu scores between the ground-truth description and LLM-generated ones range from 0.19 to 0.49.

\textbf{LLM-wise: } Without fine-tuning, DeepSeek-R1 achieved the best average (24.36\%), followed by qwen2.5-coder-instruct-32B (19.05\%).  

\textbf{{Model Size:}} Larger LLMs generally perform better than smaller LLMs. For example, llama3.1-8B only achieved 1.43\% in generating TLA+ segments, while llama3.1-70B boosts to 24.29\% in the same task. 
However, there are several exceptions worth noticing, especially for ProofInfill and Code2Proof. For example, llama-3.1-\textbf{8}B achieved 57.14\% in Code2Proof (ACSL), yet the performance drops to 21.43\% using the llama-3.1-\textbf{70}B model. 
The decrease in performance is inherently due to the fine-tuning strategy of these instruction models: they are trained to {{excel in generating rather than filling in the blanks}}~\cite{fried2022incoder}. Also, we conducted a more detailed examination of generated segments and observed that larger LLMs tend to fill in the proof segments that \textbf{\textit{not only look more plausibly correct and well-organized but also include extra content}}. The additional content, yet, is either redundant, as it repeats information that appears in the subsequent proof, or is incomplete. 
Promisingly, recent model developers have noticed such conundrums and refined their fine-tuning strategy for fill-in-the-middle tasks~\cite{deepseek-coder}.

\input{TableTex/zero-few-shot}

\input{TableTex/general-task}

\subsection{RQ2. Formal Specification Languages-wise Capability}\label{sec:rq2}

Table~\ref{tab:fewshot} shows the LLMs' performance across formal specification languages in the task of generating proof segments (SegGen). This task accounts for the most instructions and serves as the basic capability for other proof generation tasks. We can see that LLMs perform the best in ACSL (average: 34.92\%), followed by Dafny (15.97\%) while performing unsatisfactorily in other formal specification languages. The observation is reasonable because the syntax of ACSL is basically an annotation of C language, while Dafny shares similar grammar as C\# and Java. 
Thus, generating proof segments in ACSL and Dafny is generally easier than generating other specification languages. 

Note that though LLMs are generally good at generating syntactically correct and semantically consistent (\ie program behaviors are consistent with the annotated ACSL specifications) according to the `ACSL' column in Table~\ref{tab:pass1}, the specifications are usually insufficient for the verification according to the `ACSL-WP' column in Table~\ref{tab:pass1}. The results in the two ACSL columns indicate that there is a large room to improve for generating not only syntactically/semantically correct specifications, but also that the generated specifications need to be sufficient to verify the properties under verification. 

In addition, we explore whether increasing the attempts (1 $\rightarrow$ 5) with a higher temperature (0.0 $\rightarrow$ 0.2) and in-context learning could bring about improvement. The improvement ratios are shown in red in Table~\ref{tab:fewshot}. The results of Pass@5 are better than those of Pass@1, with an average score increase from 10.82\% (Dafny) to 63.64\% (ACSL) in different languages. Moreover, when using 3-shot, the performance increases dramatically, with 51.33\% (Dafny) to over five times (ACSL) improvement compared with zero-shot Pass@5. The results indicate {{the potential of in-context learning in generating correct specification languages.}}


\subsection{RQ3. Improvement by Fine-tuning}\label{sec:rq3}

We further investigate whether \fma could bring about improvement.
The lower part of Table~\ref{tab:pass1} and Table~\ref{tab:reqsplitter} shows the results. From Table~\ref{tab:pass1}, dramatic improvements 
can be observed in generating full and segmental proofs after fine-tuning. Note that the model size of fine-tuned models is 7B $\sim$ 8B, while the performance largely outperforms the 70B+ models without fine-tuning. Furthermore, after fine-tuning with formal data, \textbf{\textit{the 7 $\sim$ 8B fine-tuned models can achieve comparable or slightly better performance than Deepseek-R1-671B}}, with 27.31\% achieved by qwen2.5-coder-7B fine-tuned with \fma (R1-671B: 27.11\%).
It may suggest the possibility of distilling domain-specific small models for handier usage. 

\textbf{Task-wise:} Improvements in generation tasks (\ie ProofGen, SegGen, and ProofComplete) are substantial. 
ProofGen doubles the performance, and SegGen more than triples. The dramatic increases happen in all models fine-tuned with \fma in SegGen Task, from nearly all zeros to 29.98\% $\sim$ 90.48\%. 
An increase of 41\% can also be observed in Table~\ref{tab:reqsplitter}. {{The experimental improvements make evident the effectiveness of fine-tuning in formal verification tasks.}}

Yet, drops can be observed in fill-in-the-middle tasks (\ie ProofInfill and Code2Proof). The results echo the observation made in RQ1 (Section~\ref{sec:rq1}), where the large LLMs perform worse than small LLMs in fill-in-the-middle tasks. The results also indicate the necessity of adopting different fine-tuning strategies other than instruction tuning only.

\textbf{Fine-tuning Datasets:} Take a closer look at the LLMs fine-tuned with {{general-purpose datasets}} (\ie llama3.1-ultrachat and llama3.1-tulu) in Table~\ref{tab:pass1}, with them only, {{no or opposite effects can be observed}}. The results indicate the \textbf{\textit{complementarity}} of \fma and existing general-purpose fine-tuning datasets. Additionally, by combining with general-purpose datasets, the performance can be \textbf{\textit{further improved}} (\eg llama3.1-tulu-fma). 

\textbf{Comparison with Few-shot:} Compared with the best results in Table~\ref{tab:fewshot} achieved by 3-shot, the results after fine-tuning (Table~\ref{tab:pass1}) still generally outperform the 3-shot results. The results indicate that although in-context learning can improve LLMs' performance, the enhancement is limited. \textbf{\textit{Further significant improvements still require fine-tuning with formal data}}. This may also suggest that in-context learning alone cannot adequately address capability deficits in formal verification tasks but rather stem from a lack of knowledge.

\subsection{RQ4. Capability Migration from Formal Verification to Related Tasks}

Finally, we explore whether fine-tuning with \fma could benefit related capabilities. 
Table~\ref{tab:general-task} shows the results. The base model is llama3.1-8B, fine-tuned under two base fine-tuned datasets with and without \fma. On average, with \fma, an increase of 1.37\% to 5.15\% can be observed. Interestingly, a dramatic increase (62.53\%) can be observed in HumanEval compared with the performance of the model that is only fine-tuned with UltraChat. 
The experiment may indicate that feeding more formal data may improve LLMs' coding, reasoning, and math capabilities.

%% file: TableTex/reqsplitter.tex
\begin{table}[t!]
    \centering
    \renewcommand\arraystretch{1.2}
    \resizebox{1.0\linewidth}{!}{
\begin{tabular}{lrlllll}
\toprule
\multicolumn{1}{c}{}                                & \multicolumn{1}{c}{}                                & \multicolumn{5}{c}{\textbf{ReqSplitter}}                                                                                                                 \\
\multicolumn{1}{c}{\multirow{-2}{*}{\textbf{LLMs}}} & \multicolumn{1}{c}{\multirow{-2}{*}{\textbf{Size}}} & \multicolumn{1}{c}{TLA}      & \multicolumn{1}{c}{Coq}      & \multicolumn{1}{c}{Lean}     & \multicolumn{1}{c}{Dafny}    & \multicolumn{1}{c}{ACSL}     \\ \midrule
\multicolumn{7}{c}{\textbf{w/o fine-tune}}                                                                                                                                                                                                                           \\ \midrule
\multicolumn{1}{l|}{llama3.1-instruct}              & 8B                                                  & \cellcolor[HTML]{F9DEDF}0.33 & \cellcolor[HTML]{FBE9EA}0.26 & \cellcolor[HTML]{F8DADC}0.35 & \cellcolor[HTML]{F8D9DA}0.36 & \cellcolor[HTML]{FAE7E7}0.28 \\
\multicolumn{1}{l|}{llama3.1-instruct}              & 70B                                                 & \cellcolor[HTML]{F9DEDF}0.33 & \cellcolor[HTML]{FAE7E8}0.28 & \cellcolor[HTML]{F8DADB}0.35 & \cellcolor[HTML]{F6D0D2}0.41 & \cellcolor[HTML]{FAE3E4}0.30 \\
\multicolumn{1}{l|}{qwen2.5-instruct}               & 7B                                                  & \cellcolor[HTML]{FAE5E6}0.29 & \cellcolor[HTML]{FBEBEC}0.25 & \cellcolor[HTML]{F9DFE0}0.32 & \cellcolor[HTML]{F9DFE0}0.33 & \cellcolor[HTML]{FEF6F6}0.19 \\
\multicolumn{1}{l|}{qwen2.5-instruct}               & 72B                                                 & \cellcolor[HTML]{F9DEDF}0.33 & \cellcolor[HTML]{FAE6E7}0.28 & \cellcolor[HTML]{F9DEDF}0.33 & \cellcolor[HTML]{F8DBDC}0.34 & \cellcolor[HTML]{FCF1F1}0.22 \\
\multicolumn{1}{l|}{qwen2.5-coder-instruct}         & 7B                                                  & \cellcolor[HTML]{FAE4E5}0.29 & \cellcolor[HTML]{FBEAEA}0.26 & \cellcolor[HTML]{F8DADC}0.35 & \cellcolor[HTML]{F6CFD1}0.42 & \cellcolor[HTML]{FBECEC}0.25 \\
\multicolumn{1}{l|}{qwen2.5-coder-instruct}         & 32B                                                 & \cellcolor[HTML]{F9E1E2}0.31 & \cellcolor[HTML]{FBE9E9}0.27 & \cellcolor[HTML]{F9DEDF}0.33 & \cellcolor[HTML]{F6D3D4}0.39 & \cellcolor[HTML]{FCEFF0}0.23 \\
\multicolumn{1}{l|}{star-coder-instruct}            & 15B                                                 & \cellcolor[HTML]{FDF3F3}0.21 & \cellcolor[HTML]{FCEEEE}0.24 & \cellcolor[HTML]{FAE4E5}0.29 & \cellcolor[HTML]{FAE4E4}0.30 & \cellcolor[HTML]{FAE4E5}0.29 \\
\multicolumn{1}{l|}{deepseek-coder-instruct}        & 7B                                                  & \cellcolor[HTML]{F8DDDE}0.34 & \cellcolor[HTML]{FAE3E4}0.30 & \cellcolor[HTML]{F7D4D6}0.39 & \cellcolor[HTML]{F3C2C4}0.49 & \cellcolor[HTML]{F8D9DA}0.36 \\
\multicolumn{1}{l|}{deepseek-coder-instruct}        & 33B                                                 & \cellcolor[HTML]{F7D8D9}0.37 & \cellcolor[HTML]{FAE4E5}0.29 & \cellcolor[HTML]{F4C7C8}0.47 & \cellcolor[HTML]{F4C7C8}0.47 & \cellcolor[HTML]{F5CDCE}0.43 \\
\multicolumn{1}{l|}{deepseek-r1}                    & 671B                                                & \cellcolor[HTML]{F9DFE0}0.32 & \cellcolor[HTML]{FBE7E8}0.27 & \cellcolor[HTML]{F8DDDE}0.33 & \cellcolor[HTML]{FBE8E9}0.27 & \cellcolor[HTML]{FAE6E7}0.28 \\ \midrule
\multicolumn{2}{c}{Average}                                                                              & \multicolumn{5}{c}{31.85}                                                                                                      \\ \midrule
\multicolumn{7}{c}{w/ fine-tune}                                                                                                                                                                                                                                     \\ \midrule
\multicolumn{1}{l|}{llama3.1-\textbf{fma}}                   & 8B                                                  & \cellcolor[HTML]{F8DDDE}0.33 & \cellcolor[HTML]{FEF7F7}0.18 & \cellcolor[HTML]{F3C3C5}0.49 & \cellcolor[HTML]{F3C3C5}0.49 & \cellcolor[HTML]{F5CED0}0.42 \\
\multicolumn{1}{l|}{llama3.1-ultrachat}             & 8B                                                  & \cellcolor[HTML]{F8DADB}0.35 & \cellcolor[HTML]{FAE3E3}0.30 & \cellcolor[HTML]{F5CCCD}0.44 & \cellcolor[HTML]{F6D1D3}0.40 & \cellcolor[HTML]{F4C7C9}0.46 \\
\multicolumn{1}{l|}{llama3.1-ultrachat-\textbf{fma}}          & 8B                                                  & \cellcolor[HTML]{F7D7D8}0.37 & \cellcolor[HTML]{FAE2E3}0.30 & \cellcolor[HTML]{F1B7B9}0.56 & \cellcolor[HTML]{F0B5B8}0.57 & \cellcolor[HTML]{EEACAF}0.62 \\
\multicolumn{1}{l|}{llama3.1-tulu}                  & 8B                                                  & \cellcolor[HTML]{F6D0D2}0.41 & \cellcolor[HTML]{FAE3E4}0.30 & \cellcolor[HTML]{F5CCCE}0.43 & \cellcolor[HTML]{F6CFD0}0.42 & \cellcolor[HTML]{F6D4D5}0.39 \\
\multicolumn{1}{l|}{llama3.1-tulu-\textbf{fma}}               & 8B                                                  & \cellcolor[HTML]{F0B6B8}0.57 & \cellcolor[HTML]{F9DEDF}0.33 & \cellcolor[HTML]{F0B5B7}0.57 & \cellcolor[HTML]{F0B2B4}0.59 & \cellcolor[HTML]{EB9DA0}0.71 \\
\multicolumn{1}{l|}{qwen2.5-\textbf{fma}}                    & 7B                                                  & \cellcolor[HTML]{F9DEDF}0.33 & \cellcolor[HTML]{FFFFFF}0.13 & \cellcolor[HTML]{F2BCBE}0.53 & \cellcolor[HTML]{F0B3B5}0.58 & \cellcolor[HTML]{F7D5D7}0.38 \\
\multicolumn{1}{l|}{qwen2.5-coder-\textbf{fma}}              & 7B                                                  & \cellcolor[HTML]{F2BFC0}0.51 & \cellcolor[HTML]{FDF2F2}0.21 & \cellcolor[HTML]{F0B6B8}0.57 & \cellcolor[HTML]{EEA9AB}0.64 & \cellcolor[HTML]{EA999C}0.73 \\
\multicolumn{1}{l|}{deepseek-coder-\textbf{fma}}             & 7B                                                  & \cellcolor[HTML]{F5CCCE}0.43 & \cellcolor[HTML]{FCEDEE}0.24 & \cellcolor[HTML]{F4C7C9}0.46 & \cellcolor[HTML]{F1BABC}0.54 & \cellcolor[HTML]{F5CECF}0.43 \\ \midrule
\multicolumn{2}{c}{Average}                                                                              & \multicolumn{5}{c}{44.36 \textcolor{purple}{(\textbf{\textcolor{purple}{39\% $\uparrow$}})}}\\
\bottomrule
\multicolumn{7}{l}{* \textbf{-fma}: fine-tuned with \fma.} 
\end{tabular} 
    }
    \caption{Evaluation on Requirement Analysis}
    \label{tab:reqsplitter}
\end{table}

%% file: TableTex/zero-few-shot.tex
\begin{table*}[ht]
    \centering
    \renewcommand\arraystretch{1.2}
    \resizebox{1\textwidth}{!}{
\begin{tabular}{lrlll|lll|lll|lll|lll|lll}
\toprule
\multicolumn{1}{c|}{}                                & \multicolumn{1}{c|}{}                                & \multicolumn{3}{c|}{\textbf{TLA}}                                                                                 & \multicolumn{3}{c|}{\textbf{Coq}}                                                                                 & \multicolumn{3}{c|}{\textbf{Lean}}                                                                                & \multicolumn{3}{c|}{\textbf{Dafny}}                                                                               & \multicolumn{3}{c|}{\textbf{ACSL}}                                                                                & \multicolumn{3}{c}{\textbf{ACSL-WP}}                                                                             \\ 
\cline{3-20} 
\multicolumn{1}{c|}{}                                & \multicolumn{1}{c|}{}                                & \multicolumn{2}{c|}{\textbf{Zero-shot}}                                    & \textbf{Few-Shot}                    & \multicolumn{2}{c|}{\textbf{Zero-shot}}                                    & \textbf{Few-Shot}                    & \multicolumn{2}{c|}{\textbf{Zero-shot}}                                    & \textbf{Few-Shot}                    & \multicolumn{2}{c|}{\textbf{Zero-shot}}                                    & \textbf{Few-Shot}                    & \multicolumn{2}{c|}{\textbf{Zero-shot}}                               & \textbf{Few-Shot}                    & \multicolumn{2}{c|}{\textbf{Zero-shot}}                                    & \textbf{Few-Shot}                   \\

\multicolumn{1}{c|}{\multirow{-3}{*}{\textbf{LLMs}}} & \multicolumn{1}{c|}{\multirow{-3}{*}{\textbf{Size}}} & \multicolumn{1}{c}{\textbf{P@1}} & \multicolumn{1}{c|}{\textbf{P@5}} & \multicolumn{1}{c|}{\textbf{P@1}} & \multicolumn{1}{c}{\textbf{P@1}} & \multicolumn{1}{c|}{\textbf{P@5}} & \multicolumn{1}{c|}{\textbf{P@1}} & \multicolumn{1}{c}{\textbf{P@1}} & \multicolumn{1}{c|}{\textbf{P@5}} & \multicolumn{1}{c|}{\textbf{P@1}} & \multicolumn{1}{c}{\textbf{P@1}} & \multicolumn{1}{c|}{\textbf{P@5}} & \multicolumn{1}{c|}{\textbf{P@1}} & \multicolumn{1}{c}{\textbf{P@1}} & \multicolumn{1}{c|}{\textbf{P@5}} & \multicolumn{1}{c|}{\textbf{P@1}} & \multicolumn{1}{c}{\textbf{P@1}} & \multicolumn{1}{c|}{\textbf{P@5}} & \multicolumn{1}{c}{\textbf{P@1}} \\ \midrule
\multicolumn{1}{l|}{llama3.1-instruct}               & 8B                                                   & \cellcolor[HTML]{FFFFFF}1.43        & \cellcolor[HTML]{F9FCFB}4.29         & \cellcolor[HTML]{E1EDED}11.43        & \cellcolor[HTML]{FCFDFD}1.19        & \cellcolor[HTML]{FBFDFD}1.90         & \cellcolor[HTML]{CDE1E1}10.12        & \cellcolor[HTML]{FFFFFF}6.06        & \cellcolor[HTML]{EDF4F4}8.33         & \cellcolor[HTML]{E0EDED}9.85         & \cellcolor[HTML]{F0F6F6}8.33        & \cellcolor[HTML]{E7F1F0}11.11        & \cellcolor[HTML]{C1DADA}19.44        & \cellcolor[HTML]{FFFFFF}0.00        & \cellcolor[HTML]{FCFDFD}3.17         & \cellcolor[HTML]{D3E5E5}38.10        & \cellcolor[HTML]{FFFFFF}0.00        & \cellcolor[HTML]{FFFFFF}0.00         & \cellcolor[HTML]{AECFCE}12.70       \\
\multicolumn{1}{l|}{llama3.1-instruct}               & 70B                                                  & \cellcolor[HTML]{8FBCBB}24.29       & \cellcolor[HTML]{8FBCBB}27.14        & \cellcolor[HTML]{8FBCBB}38.57        & \cellcolor[HTML]{8FBCBB}4.86        & \cellcolor[HTML]{E4EFEF}6.01         & \cellcolor[HTML]{F2F7F7}3.46         & \cellcolor[HTML]{8FBCBB}13.64       & \cellcolor[HTML]{A2C8C7}17.42        & \cellcolor[HTML]{AFCFCE}15.91        & \cellcolor[HTML]{E0ECEC}11.11       & \cellcolor[HTML]{DAE9E9}13.89        & \cellcolor[HTML]{CEE2E1}16.67        & \cellcolor[HTML]{8FBCBB}39.68       & \cellcolor[HTML]{C3DBDB}52.38        & \cellcolor[HTML]{A7CBCA}76.19        & \cellcolor[HTML]{B5D3D2}3.17        & \cellcolor[HTML]{E1EDED}4.76         & \cellcolor[HTML]{A4C9C8}14.29       \\
\multicolumn{1}{l|}{qwen2.5-instruct}                & 7B                                                   & \cellcolor[HTML]{FFFFFF}1.43        & \cellcolor[HTML]{FFFFFF}2.86         & \cellcolor[HTML]{F3F8F8}5.71         & \cellcolor[HTML]{FFFFFF}1.05        & \cellcolor[HTML]{FEFEFE}1.39         & \cellcolor[HTML]{BED8D8}12.80        & \cellcolor[HTML]{E9F2F2}7.58        & \cellcolor[HTML]{E7F1F0}9.09         & \cellcolor[HTML]{E0EDED}9.85         & \cellcolor[HTML]{FFFFFF}5.56        & \cellcolor[HTML]{FFFFFF}5.56         & \cellcolor[HTML]{DAE9E9}13.89        & \cellcolor[HTML]{FBFDFD}1.59        & \cellcolor[HTML]{FCFDFD}3.17         & \cellcolor[HTML]{B2D1D1}66.67        & \cellcolor[HTML]{DAE9E9}1.59        & \cellcolor[HTML]{EBF3F3}3.17         & \cellcolor[HTML]{9AC3C2}15.87       \\
\multicolumn{1}{l|}{qwen2.5-instruct}                & 72B                                                  & \cellcolor[HTML]{C7DEDD}12.86       & \cellcolor[HTML]{BED8D8}17.14        & \cellcolor[HTML]{BBD6D6}24.29        & \cellcolor[HTML]{BDD8D7}3.29        & \cellcolor[HTML]{F0F6F6}3.80         & \cellcolor[HTML]{8FBCBB}21.05        & \cellcolor[HTML]{A6CAC9}12.12       & \cellcolor[HTML]{BBD7D6}14.39        & \cellcolor[HTML]{8FBCBB}19.70        & \cellcolor[HTML]{8FBCBB}25.00       & \cellcolor[HTML]{A8CBCB}25.00        & \cellcolor[HTML]{A8CBCB}25.00        & \cellcolor[HTML]{DCEAEA}12.70       & \cellcolor[HTML]{EDF5F4}15.87        & \cellcolor[HTML]{97C1C0}90.48        & \cellcolor[HTML]{8FBCBB}4.76        & \cellcolor[HTML]{E1EDED}4.76         & \cellcolor[HTML]{9AC3C2}15.87       \\
\multicolumn{1}{l|}{qwen2.5-coder-instruct}          & 7B                                                   & \cellcolor[HTML]{F8FBFB}2.86        & \cellcolor[HTML]{F2F8F8}5.71         & \cellcolor[HTML]{DDEBEB}12.86        & \cellcolor[HTML]{DFECEC}2.14        & \cellcolor[HTML]{F5F9F9}2.99         & \cellcolor[HTML]{B5D3D2}14.30        & \cellcolor[HTML]{A6CAC9}12.12       & \cellcolor[HTML]{C1DADA}13.64        & \cellcolor[HTML]{96C0BF}18.94        & \cellcolor[HTML]{E0ECEC}11.11       & \cellcolor[HTML]{E7F1F0}11.11        & \cellcolor[HTML]{C1DADA}19.44        & \cellcolor[HTML]{F7FAFA}3.17        & \cellcolor[HTML]{FAFCFC}4.76         & \cellcolor[HTML]{9BC3C2}87.30        & \cellcolor[HTML]{FFFFFF}0.00        & \cellcolor[HTML]{FFFFFF}0.00         & \cellcolor[HTML]{8FBCBB}17.46       \\
\multicolumn{1}{l|}{qwen2.5-coder-instruct}          & 32B                                                  & \cellcolor[HTML]{C7DEDD}12.86       & \cellcolor[HTML]{C4DCDB}15.71        & \cellcolor[HTML]{A9CCCB}30.00        & \cellcolor[HTML]{A6CAC9}4.07        & \cellcolor[HTML]{EAF2F2}4.96         & \cellcolor[HTML]{9EC5C5}18.40        & \cellcolor[HTML]{C7DEDD}9.85        & \cellcolor[HTML]{B5D3D2}15.15        & \cellcolor[HTML]{C1DADA}13.64        & \cellcolor[HTML]{8FBCBB}25.00       & \cellcolor[HTML]{9CC4C3}27.78        & \cellcolor[HTML]{8FBCBB}30.56        & \cellcolor[HTML]{B3D2D1}26.98       & \cellcolor[HTML]{D9E8E8}33.33        & \cellcolor[HTML]{8FBCBB}96.83        & \cellcolor[HTML]{8FBCBB}4.76        & \cellcolor[HTML]{E1EDED}4.76         & \cellcolor[HTML]{8FBCBB}17.46       \\
\multicolumn{1}{l|}{deepseek-coder-instruct}         & 6.7B                                                 & \cellcolor[HTML]{8FBCBB}4.29        & \cellcolor[HTML]{8FBCBB}8.57         & \cellcolor[HTML]{F2F8F8}4.29         & \cellcolor[HTML]{FFFFFF}1.73        & \cellcolor[HTML]{FFFFFF}2.82         & \cellcolor[HTML]{E4EFEF}8.79         & \cellcolor[HTML]{FFFFFF}3.79        & \cellcolor[HTML]{FFFFFF}5.30         & \cellcolor[HTML]{D6E7E7}17.42        & \cellcolor[HTML]{FFFFFF}5.56        & \cellcolor[HTML]{FFFFFF}5.56         & \cellcolor[HTML]{D4E6E6}19.44        & \cellcolor[HTML]{FFFFFF}0.00        & \cellcolor[HTML]{FFFFFF}4.76         & \cellcolor[HTML]{B0D0D0}60.32        & \cellcolor[HTML]{8FBCBB}0.00        & \cellcolor[HTML]{FFFFFF}0.00         & \cellcolor[HTML]{BFD9D9}9.52        \\
\multicolumn{1}{l|}{deepseek-coder-instruct}         & 33B                                                  & \cellcolor[HTML]{FFFFFF}2.86        & \cellcolor[HTML]{FFFFFF}2.86         & \cellcolor[HTML]{D5E7E7}18.57        & \cellcolor[HTML]{8FBCBB}1.94        & \cellcolor[HTML]{8FBCBB}3.23         & \cellcolor[HTML]{DCEBEB}11.27        & \cellcolor[HTML]{8FBCBB}4.55        & \cellcolor[HTML]{8FBCBB}6.82         & \cellcolor[HTML]{D1E4E4}22.73        & \cellcolor[HTML]{8FBCBB}11.11       & \cellcolor[HTML]{8FBCBB}13.89        & \cellcolor[HTML]{D2E5E5}22.22        & \cellcolor[HTML]{8FBCBB}3.17        & \cellcolor[HTML]{8FBCBB}25.40        & \cellcolor[HTML]{94BFBE}92.06        & \cellcolor[HTML]{8FBCBB}0.00        & \cellcolor[HTML]{8FBCBB}1.59         & \cellcolor[HTML]{99C2C1}15.87       \\ \midrule
\multicolumn{2}{r|}{Language-wise Average}                                                                  & \multicolumn{3}{c|}{12.20}                                                                                        & \multicolumn{3}{c|}{6.15}                                                                                         & \multicolumn{3}{c|}{11.99}                                                                                        & \multicolumn{3}{c|}{15.97}                                                                                        & \multicolumn{3}{c|}{34.92}                                                                                        & \multicolumn{3}{c}{6.35}                                                                                         \\ \midrule
\multicolumn{2}{r|}{Average}                                                                                
& 7.86                                & 10.54                                & 18.21                                & 2.53                                & 3.39                                 & 12.53                                & 8.71                                & 11.27                                & 16.00                                & 12.85                               & 14.24                                & 20.83                                & 10.91                               & 17.86                                & 75.99                                & 1.79                                & 2.38                                 & 14.88                               \\
\multicolumn{2}{r|}{Improvement Ratio}                                                                            &                                     & \textcolor{purple}{34.09\%$\uparrow$}                              & \textcolor{purple}{72.88\%$\uparrow$}                              &                                     & \textcolor{purple}{33.67\%$\uparrow$}                              & \textcolor{purple}{269.80\%$\uparrow$}                             &                                     & \textcolor{purple}{29.35\%$\uparrow$}                              & \textcolor{purple}{42.02\%$\uparrow$}                              &                                     & \textcolor{purple}{10.81\%$\uparrow$}                              & \textcolor{purple}{46.34\%$\uparrow$}                              &                                     & \textcolor{purple}{63.64\%$\uparrow$}                              & \textcolor{purple}{325.56\%$\uparrow$}                             &                                     & \textcolor{purple}{33.33\%$\uparrow$}                              & \textcolor{purple}{525.00\%$\uparrow$}                           
\\
\bottomrule
\multicolumn{20}{l}{* The improvement ratios shown in ``Pass@5'' column are calculated by comparing with the scores in Pass@1, and the ratios shown in ``Few-shot'' column are calculated by comparing with the scores in Pass@5. }\\
\multicolumn{20}{l}{\textbf{* ACSL and ACSL-WP}: `ACSL' uses `frama-c' tool for syntax and basic ACSL checking; `ACSL-WP' uses `frama-c' with the WP plugin for specification verification.}

\end{tabular}
    }
    \caption{\textbf{RQ2: Language-wise LLMs' Performance}. Pass@1 and Pass@5 Accuracy in Generating Proof Segments Across Formal Specification Languages Under Zero/Few-shot without fine-tuning.}
    \label{tab:fewshot}
\end{table*}

%% file: TableTex/general-task.tex
\begin{table*}[ht]
    \centering
    \renewcommand\arraystretch{1.2}
    \resizebox{1\textwidth}{!}{

\begin{tabular}{l|cc|cc|cc||cc||cc|cc|cc||cc}
\toprule
\multicolumn{1}{c|}{} & \multicolumn{6}{c||}{\textbf{MATH}} & \multicolumn{2}{c||}{\textbf{Reasoning}} & \multicolumn{6}{c|}{\textbf{Coding}} & \multicolumn{2}{c}{} \\ \cline{2-15}
\multicolumn{1}{l|}{\multirow{-2}{*}{\textbf{\begin{tabular}[l]{@{}l@{}}Fine-tuning \\ Dataset\end{tabular}}}} & \multicolumn{2}{c|}{MATH} & \multicolumn{2}{c|}{GSM-8K} & \multicolumn{2}{c||}{\textbf{Average}} & \multicolumn{2}{c||}{bbh} & \multicolumn{2}{c|}{HumanEval} & \multicolumn{2}{c|}{MBPP} & \multicolumn{2}{c||}{\textbf{Average}} & \multicolumn{2}{c}{\multirow{-2}{*}{\textbf{Average}}} \\
\midrule

UltraChat & \textbf{17.54} & \textbf{} & 61.33 &  & 39.44 &  & \textbf{62.64} & \textbf{} & 19.51 &  & \textbf{36.4} & \textbf{} & 27.96 &  & 39.48 &  \\
UltraChat + \textbf{fma} & 16.16 & \textcolor{teal}{7.87\%$\downarrow$} & \textbf{62.32} & \textbf{\textcolor{purple}{1.61\% $\uparrow$}} & 39.24 & \textcolor{teal}{0.49\%$\downarrow$} & 62.14 & \textcolor{teal}{0.80\%$\downarrow$} & \textbf{31.71} & \textbf{\textcolor{purple}{62.53\% $\uparrow$}} & 35.2 & {\textcolor{teal}{3.30\%$\downarrow$}} & \textbf{33.46} & \textbf{\textcolor{purple}{19.67\% $\uparrow$}} & \textbf{41.51} & \multicolumn{1}{c}{\textbf{\textcolor{purple}{5.14\% $\uparrow$}}} \\
\midrule

{tulu3} & 27.36 &  & \textbf{75.82} & \textbf{} & 51.59 &  & 62.47 &  & 64.02 &  & 48.8 &  & 56.41 &  & 55.01 &  \\
{tulu3 + \textbf{fma}} & \cellcolor[HTML]{FFFFFF}\textbf{29.48} & \textbf{\textcolor{purple}{7.75\% $\uparrow$}} & 75.44 & \textcolor{teal}{0.50\%$\downarrow$} & 52.46 & \textbf{\textcolor{purple}{1.69\% $\uparrow$}} & \textbf{63.16} & \textbf{\textcolor{purple}{1.10\% $\uparrow$}} & \textbf{64.63} & \textbf{\textcolor{purple}{0.95\% $\uparrow$}} & \textbf{49.4} & \textbf{\textcolor{purple}{1.23\% $\uparrow$}} & \textbf{57.02} & \textbf{\textcolor{purple}{1.07\% $\uparrow$}} & \textbf{55.76} & \multicolumn{1}{c}{\textbf{\textcolor{purple}{1.37\% $\uparrow$}}} \\
\bottomrule

\end{tabular}

    }
    \caption{\textbf{RQ4: Capability Migration from \fma to Math, Reasoning, and Coding}.}
    \label{tab:general-task}
\end{table*}

%% file: Tex/7-Conclusion.tex
\section{Conclusion}
This paper contributes a comprehensive assessment and formulation to understand LLMs' capability in formal verification. 
We constructed {{18k high-quality instruction-response pairs}} across five formal specification languages in six tasks. The fine-tuned models, fine-tuning data, and the benchmark are released to facilitate subsequent studies.


%% file: Tex/8-Limitation.tex
\section*{Limitations}
This paper has two primary limitations that offer avenues for future research.
First, the primary limitation of our work is that our benchmark relies on model-generated data. While this approach effectively reduces manual efforts; it may introduce biases and data leakage issues in the dataset towards the models that generated the data. To address this limitation, we use gpt-4o to generate the natural language descriptions, while during the evaluation, we use other LLMs for evaluation. 
Second, another limitation of our work lies in the validation design. When creating ProofInfill and ProofComplete data, it is possible that the properties to be verified or theorems to be proven are masked. If LLMs happened not to generate these properties/theorems, the generated ``proofs/models'' could escape the verifier/checker, mistakenly labeling the output as correct. To avoid this scenario, we include the requirement descriptions as part of the input, guiding LLMs to generate the necessary properties or theorems without omission.


%% file: Tex/9-Acknowledgements.tex
\section{Acknowledgments}
We sincerely thank the reviewers for their insightful comments and valuable suggestions.
This work was supported in part by the Hong Kong Research Grants Council/General Research Fund (HKSAR RGC/GRF, Grant No. 16206524), the National Natural Science Foundation of China (Grant Nos. 62372193, 62192734, 62302375, 62472339), CAS Project for Young Scientists in Basic Research (Grant No.YSBR-040) and the Basic Research Program of ISCAS (Grant No. ISCAS-ZD-202402).

%% file: Appendix/main-appendix.tex
\newpage

\input{Appendix/Background}

\input{Appendix/syntax-errors}

\input{Appendix/examples}

\input{Appendix/repos}

\input{Appendix/pass5}

\input{Appendix/prompt}

%% file: Appendix/Background.tex
\section{Related Work}
The formal specification datasets or benchmarks offer a standard, well-defined set of problems, providing a shared challenge that helps build a community of practice among researchers. According to different verification techniques, the existing benchmarks mainly fall into two categories; we discuss them separately.

\subsection{Theorem Proving Datasets}
Formal theorem proving represents theorems and proofs in a machine-verifiable format~\cite{cook2023complexity}, ensuring their correctness using rigorous logical rules. A recent survey~\cite{li2024survey} summarized the existing datasets for theorem proving. In particular, the \textbf{\textit{informal benchmarks}} craft the proofs from various sources such as ProofWiki, textbooks, and public corpus. NL-PS~\cite{ferreira2020NL-PS} first builds a natural language premise selection dataset source from ProofWiki. Similarly, NaturalProofs~\cite{welleck2021naturalproofsmathematicaltheoremproving} further incorporates data from Stacks and textbooks, resulting in a dataset with roughly 25k examples. 
Adapted from it, NaturalProofs-Gen~\cite{welleck2022naturalprover} contains around 14.5k theorems for informal proof generation. Moreover, MATcH~\cite{li2023bert} constructs over 180k statement-proof pairs for matching using the MREC corpus~\footnote{\url{https://mir.fi.muni.cz/MREC/}}. 





For \textbf{\textit{formal datasets}}, a line of efforts focuses on extracting and cleaning theorems and proofs written in various specification languages (\eg Coq, Isabelle, Lean) from established formal libraries and verification projects. For example, LeanDojo~\cite{yang2023leandojo} extracts over 98k theorems and proofs with 130k premises from Lean mathlib~\cite{lean-library}. Besides extracting data from existing projects, several works \textbf{\textit{manually}} annotate or formalize the problems in natural language. For example, MiniF2F~\cite{zheng2021minif2f} manually formalizes 488 Olympiad-level problems across 4 proof systems and equally splits them into a validation set and a test set. FIMO~\cite{fimo20223} and ProofNet~\cite{proofnet2023} formalize the theorem statements of the International Mathematical Olympiad and undergraduate-level problems in Lean. 
In addition, datasets for Dafny also attract research contributions because industries like Amazon adopted Dafny to verify cryptographic libraries,  authorization protocols, a random number generator, and the Ethereum virtual machine. Dafny datasets such as CloverBench~\cite{sun2024cloverclosedloopverifiablecode} and DafnyGym~\cite{mugnier2024laurelgeneratingdafnyassertions}. 



\subsection{Model checking datasets}

Model checking is an automated technique used in computer science and formal methods to verify the correctness of systems, particularly those with finite state spaces. It systematically checks whether a system's model satisfies a given specification, usually expressed in formal specification languages. The basic idea is to explore all possible system states to ensure the desired properties hold in every conceivable scenario. 

Model checking {{benchmarks}} are less than that for theorem proving. Currently, there are few model-checking benchmarks for proving, while several model-checking subjects are going with specific model-checking languages such as CMurphi~\cite{della2013cgmurphi} and TLA+~\cite{tla1999}. In particular, CMurphi is a software tool used to verify concurrent and distributed systems through explicit state enumeration. It implements the Murphi verification language, which allows users to describe finite-state systems in a procedural style. The core principle behind CMurphi is to explore the state space of a system exhaustively to check for violations of specified invariants or properties. Another example is TLA+ (Temporal Logic of Actions), a high-level language for modeling programs and systems suitable for concurrent and distributed systems. 




%% file: Appendix/syntax-errors.tex
\section{Proportion of Failures Caused by Syntax Error}\label{sec:syntax_error}
We listed the proportions of failures caused by syntax errors for each LLM and each task in Table~\ref{tab:syntax_error_breakdown}.
We used a set of pre-defined keywords (summarized in Table~\ref{tab:syntax_error_keywords} to identify if a verification failure is caused by syntax errors.
Specifically, we consider a failure caused by syntax error if its error message contains at least one keyword in Table~\ref{tab:syntax_error_keywords}.

\begin{table}[ht]
    \centering
    \resizebox{1.0\linewidth}{!}{
\begin{tabular}{lll}
\toprule
 & Language & Keywords \\ \midrule
1 & Coq &  ``Syntax Error:'' \\
2 & Lean4 &  ``unexpected token '', ``unknown identifier'', ``type mismatch''\\
3 & ACSL & ``unexpected token'' \\
4 & Dafny & ``type errors detected'' \\
5 & TLA+ & ``***Parse Error***'', ``Unknown operator''\\
\bottomrule

\end{tabular}
}
\caption{Keywords for identifying syntax error raised by each language's verifier.}
\label{tab:syntax_error_keywords}
\end{table}

\begin{table*}[th]
    \centering
    \resizebox{1\linewidth}{!}{
\begin{threeparttable}
\begin{tabular}{lrrrrrrrrrrrrrrrrrrrrr}
\toprule
\multicolumn{1}{c}{\multirow{2}{*}{\textbf{LLMs}}} & \multicolumn{1}{c}{\multirow{2}{*}{\textbf{Size}}} & \multicolumn{5}{c}{\textbf{ProofGen}}                                                                                                & \multicolumn{5}{c}{\textbf{SegGen}}                                                                                                   & \multicolumn{4}{c}{\textbf{ProofComplete}}                                                               & \multicolumn{5}{c}{\textbf{ProofInfill}}                                                                                             & \multicolumn{1}{c}{\textbf{Cd2Prf}} \\
\multicolumn{1}{c}{}                               & \multicolumn{1}{c}{}                               & \multicolumn{1}{c}{TLA} & \multicolumn{1}{c}{Coq} & \multicolumn{1}{c}{Lean} & \multicolumn{1}{c}{Dafny} & \multicolumn{1}{c}{ACSL}  & \multicolumn{1}{c}{TLA} & \multicolumn{1}{c}{Coq} & \multicolumn{1}{c}{Lean} & \multicolumn{1}{c}{Dafny} & \multicolumn{1}{c}{ACSL}   & \multicolumn{1}{c}{TLA} & \multicolumn{1}{c}{Coq} & \multicolumn{1}{c}{Lean} & \multicolumn{1}{c}{Dafny} & \multicolumn{1}{c}{TLA} & \multicolumn{1}{c}{Coq} & \multicolumn{1}{c}{Lean} & \multicolumn{1}{c}{Dafny} & \multicolumn{1}{c}{ACSL}  & \multicolumn{1}{c}{ACSL}            \\ \hline
\multicolumn{1}{l|}{llama3.1-instruct}             & \multicolumn{1}{r|}{8B}                            & 0/13                    & 18/45                   & 34/58                    & 0/10                      & \multicolumn{1}{r|}{1/14} & 43/69                   & 742/2910                & 78/124                   & 4/33                      & \multicolumn{1}{r|}{4/63}  & 4/11                    & 45/81                   & 16/34                    & \multicolumn{1}{r|}{1/5}  & 16/34                   & 44/157                  & 27/60                    & 0/13                      & \multicolumn{1}{r|}{1/14} & 0/14                                \\
\multicolumn{1}{l|}{llama3.1-instruct}             & \multicolumn{1}{r|}{70B}                           & 4/12                    & 10/44                   & 40/53                    & 1/8                       & \multicolumn{1}{r|}{1/14} & 43/53                   & 71/2802                 & 75/114                   & 16/32                     & \multicolumn{1}{r|}{5/61}  & 6/8                     & 44/81                   & 10/31                    & \multicolumn{1}{r|}{1/5}  & 15/34                   & 39/153                  & 22/49                    & 0/13                      & \multicolumn{1}{r|}{2/14} & 3/14                                \\
\multicolumn{1}{l|}{qwen2.5-instruct}              & \multicolumn{1}{r|}{7B}                            & 12/13                   & 13/46                   & 33/56                    & 1/10                      & \multicolumn{1}{r|}{0/14} & 26/69                   & 302/2914                & 53/122                   & 5/34                      & \multicolumn{1}{r|}{6/62}  & 9/11                    & 35/73                   & 8/29                     & \multicolumn{1}{r|}{1/5}  & 14/34                   & 66/156                  & 31/53                    & 0/13                      & \multicolumn{1}{r|}{4/14} & 3/14                                \\
\multicolumn{1}{l|}{qwen2.5-instruct}              & \multicolumn{1}{r|}{72B}                           & 10/12                   & 6/44                    & 28/48                    & 1/8                       & \multicolumn{1}{r|}{9/13} & 47/61                   & 40/2848                 & 54/116                   & 11/27                     & \multicolumn{1}{r|}{0/60}  & 7/11                    & 35/74                   & 8/24                     & \multicolumn{1}{r|}{1/5}  & 13/34                   & 59/149                  & 20/49                    & 0/11                      & \multicolumn{1}{r|}{0/14} & 1/14                                \\
\multicolumn{1}{l|}{qwen2.5-coder-instruct}        & \multicolumn{1}{r|}{7B}                            & 11/13                   & 18/46                   & 38/55                    & 0/10                      & \multicolumn{1}{r|}{0/14} & 60/68                   & 22/2882                 & 72/116                   & 6/32                      & \multicolumn{1}{r|}{53/63} & 10/11                   & 12/80                   & 8/32                     & \multicolumn{1}{r|}{2/6}  & 17/34                   & 34/159                  & 26/50                    & 1/12                      & \multicolumn{1}{r|}{0/14} & 0/14                                \\
\multicolumn{1}{l|}{qwen2.5-coder-instruct}        & \multicolumn{1}{r|}{32B}                           & 11/13                   & 12/44                   & 29/52                    & 1/10                      & \multicolumn{1}{r|}{2/14} & 50/61                   & 31/2825                 & 68/119                   & 9/27                      & \multicolumn{1}{r|}{25/60} & 5/8                     & 24/79                   & 6/23                     & \multicolumn{1}{r|}{3/4}  & 17/33                   & 51/152                  & 16/43                    & 0/12                      & \multicolumn{1}{r|}{0/12} & 0/12                                \\
\multicolumn{1}{l|}{deepseek-coder-instruct}       & \multicolumn{1}{r|}{7B}                            & 10/13                   & 4/46                    & 31/54                    & 1/10                      & \multicolumn{1}{r|}{0/14} & 35/67                   & 9/2894                  & 62/127                   & 3/34                      & \multicolumn{1}{r|}{43/63} & 9/10                    & 14/82                   & 11/31                    & \multicolumn{1}{r|}{2/5}  & 22/34                   & 18/160                  & 23/56                    & 0/10                      & \multicolumn{1}{r|}{1/14} & 1/14                                \\
\multicolumn{1}{l|}{deepseek-coder-instruct}       & \multicolumn{1}{r|}{33B}                           & 12/13                   & 0/47                    & 46/58                    & 4/10                      & \multicolumn{1}{r|}{0/14} & 66/68                   & 52/2888                 & 81/126                   & 8/32                      & \multicolumn{1}{r|}{45/63} & 10/11                   & 3/81                    & 6/31                     & \multicolumn{1}{r|}{2/4}  & 19/34                   & 6/161                   & 23/50                    & 0/14                      & \multicolumn{1}{r|}{0/14} & 1/14                                \\
\multicolumn{1}{l|}{star-coder-instruct}           & \multicolumn{1}{r|}{15B}                           & 8/12                    & 0/46                    & 27/54                    & 2/7                       & \multicolumn{1}{r|}{3/14} & 36/51                   & 10/2887                 & 67/124                   & 12/26                     & \multicolumn{1}{r|}{1/63}  & 6/8                     & 4/74                    & 7/27                     & \multicolumn{1}{r|}{1/4}  & 20/34                   & 6/149                   & 23/49                    & 0/12                      & \multicolumn{1}{r|}{0/14} & 1/14                                \\
\bottomrule
\end{tabular}
\begin{tablenotes}
\item [1] Denominator represents the total number of failures.
\end{tablenotes}
\end{threeparttable}
}
\caption{Proportion of Verification Failures Caused by Syntax Errors}
\label{tab:syntax_error_breakdown}
\end{table*}

%% file: Appendix/examples.tex
\section{Example Specifications in \fma and \fmb}
The examples of the five formal specification languages are shown in Figure~\ref{fig:example}. 

\input{FigureTex/example}

%% file: FigureTex/example.tex
\begin{figure*}[ht]
    \centering
    \includegraphics[width=1.0\textwidth]{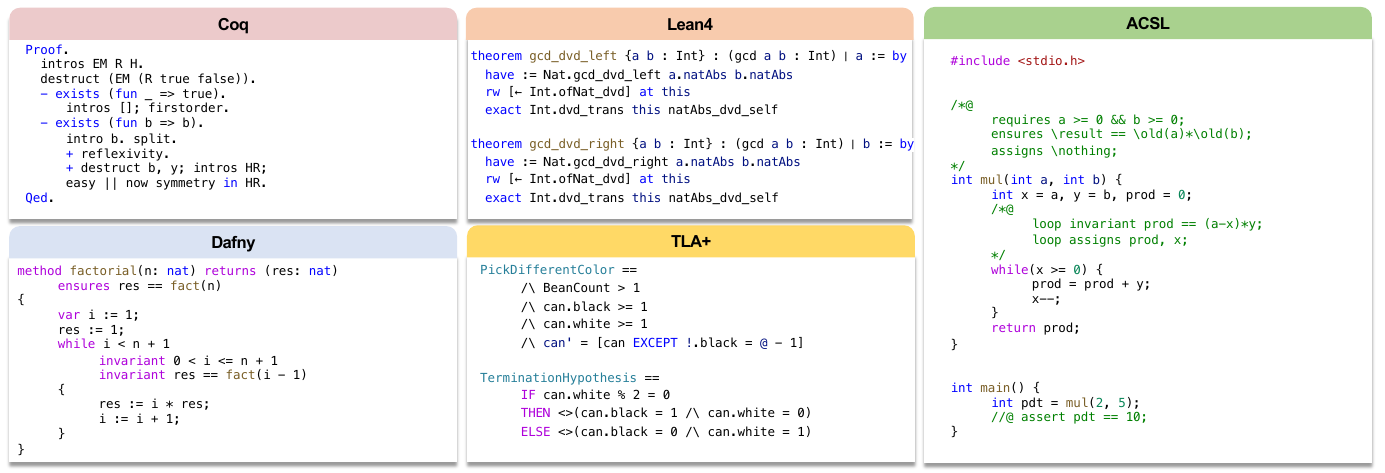}
    \setlength{\abovecaptionskip}{-0pt}
    \setlength{\belowcaptionskip}{-0pt}
    \caption{Formal Specification Languages in FM-bench}
    \label{fig:example}
\end{figure*}

%% file: Appendix/repos.tex
\section{Collected Repositories}\label{sec:repo}
We listed the repositories that were collected for data construction in the following. Note that one can easily add more repositories into \fma and \fmb.

\noindent For ACSL: 
\begin{itemize}
    \item \url{https://github.com/manavpatnaik/frama-c-problems}
    \item \url{https://github.com/fraunhoferfokus/acsl-by-example}
\end{itemize}

\noindent For TLA+:
\begin{itemize}
    \item \url{https://github.com/tlaplus/Examples}
\end{itemize}

\noindent For Lean4:
\begin{itemize}
    \item \url{https://github.com/leanprover/lean4}
\end{itemize}

\noindent For Coq:
\begin{itemize}
    \item \url{https://github.com/coq/coq}
\end{itemize}

\noindent For Dafny:
\begin{itemize}
    \item \url{https://github.com/vladstejeroiu/Dafny-programs}
\end{itemize}

%% file: Appendix/pass5.tex
\section{Complete Evaluation Result}

The Pass@1 and Pass@5 are shown in Table~\ref{tab:pass1-5}. It is a completed version of Table~\ref{tab:pass1}.

\input{TableTex/pass5}

%% file: TableTex/pass5.tex
\begin{table*}[ht]
    \centering
    \resizebox{1\linewidth}{!}{
    
\begin{tabular}{lrllllllllllllllllllllllll}
\toprule
\multicolumn{26}{c}{\textbf{Pass @ 1}}                                                                                                                                                                                                                                                                                                                                                                                                                                                                                                                                                                                                                                                                                                                                                                                                                                                                    \\ \midrule
\multicolumn{1}{c}{}                                & \multicolumn{1}{c}{}                                & \multicolumn{6}{c}{\textbf{ProofGen}}                                                                                                                                                         & \multicolumn{6}{c}{\textbf{SegGen}}                                                                                                                                                           & \multicolumn{4}{c}{\textbf{ProofComplete}}                                                                                    & \multicolumn{6}{c}{\textbf{ProofInfill}}                                                                                                                                                      & \multicolumn{2}{c}{\textbf{Cd2Prf}}                           \\
\bottomrule
\end{tabular}

    }
    \caption{Full Experiment Results on Pass@1 and Pass@5}
    \label{tab:pass1-5}
\end{table*}

%% file: Appendix/prompt.tex
\section{Prompt Design}

We listed the prompts that are used for data preparation and inference in the following. For \textbf{data preparation}, as shown in Figure~\ref{prompt:gen}, to generate descriptions for the given proof segments, the prompt template consists of five parts: (1) Role description, (2) Domain knowledge of TLA+, (3) Task description, (4) Few-shot examples (we show one example in the figure, while three-shots were used in RQ2), (5) The proof or proof segment to be summarized. 

For the \textbf{inference}, for each task, we designed five different instructions to avoid overfitting. The prompts for each task are shown in Figure~\ref{prompt:seggen} $\sim$ Figure~\ref{prompt:code2proof}. For each task, we first randomly choose one instruction and concat the inputs.

\input{Appendix/prompts/generate}

\input{Appendix/prompts/inference}

%% file: Appendix/prompts/generate.tex
\begin{figure*}[h]
\begin{tcolorbox}[colframe=cyan!40!black, title=\textbf{Prompt for Description Generation. Take TLA+ as an example.}]

\# \textbf{Role description}

As an expert in TLA+, you are good at understanding and writing TLA+.\
TLA+ is a formal specification language used for modeling and verifying concurrent and distributed systems.\\

\# \textbf{Domain knowledge}

1. The logical operators supported by TLA+ include: 
$/\backslash$ (and), $\backslash/$ (or), $\sim$ (not), $=>$ (Implication), $<=>$ (Bidirectional implication), \texttt{TRUE}, \texttt{FALSE},  $\backslash$A (Universal Quantification), $\backslash$E (Existential Quantification) 

2. The sets operators supported by TLA+ include: 
= (Equality), \# (not equal), $\backslash$union (Union), $\backslash$intersect (Intersection), $\backslash$in (Membership), $\backslash$notin (Not in), $\backslash$subseteq (Subset Equal), $\backslash$(Difference).

3. The temporal Operators supported by TLA+ include \texttt{[] x > 0}, which is an example of [] (Always). It means that at all times, the value of variable x is greater than 0. \texttt{<> x = 0} is an example of <> (Eventually). It means that at some point in time, the value of variable x becomes 0. 

4. Built-in keywords and operators in TLA+ include:
`MODULE, `EXTENDS', `CONSTANT', `INSTANCE', `VARIABLE', `ASSUME',
`PROVE', `INIT', `NEXT', `ACTION', `SPECIFICATION', `IF', `ELSE', `WITH', 
`CASE', `THEN', `LET', `IN', `CHOOSE', `ENABLED', `UNCHANGED', `DOMAIN'. \\

\# \textbf{Task description}

Given a TLA+ code snippet, you need to summarize the given TLA+ in several sentences in detail.  \\

\# \textbf{Example Input and Output}\\
\#\# \textbf{Code}\\
Return(c,S) == \\
  $/\backslash$ S \# \{\} $/\backslash$ S $\backslash$subseteq alloc[c] \\
  $/\backslash$ alloc' = [alloc EXCEPT ![c] = \@ $\backslash$ S]\\
  $/\backslash$ UNCHANGED unsat 

\#\# \textbf{Description}:\\
An operation `Return(c,S)' that represents the return of a set of resources by a client. It satisfies the following conditions:\\
  - The set `S' is not empty and `S' must be a subset of the set of allocated resources to the client `c'.\\
  - The `alloc' is updated by assigning the difference between the current set of allocated resources and the set `S' to the `c' index of `alloc' except `c'. \\
  - The `unsat' remains unchanged. \\
  
\# \textbf{Code to be described:}\\
<A proof segment>

\end{tcolorbox}
\caption{Prompt for generating TLA+ description. The prompt templates for other formal specification languages are in the same structure.}
\label{prompt:gen}
\end{figure*}

%% file: Appendix/prompts/inference.tex
\begin{figure*}[th]
\begin{tcolorbox}[colframe=cyan!40!black, title=\textbf{Prompt for SegGen Task (lang: a placeholder to be replaced by each formal specification language name.)}]

\# \textbf{Task Description (SegGen)}

1. Translate the given natural language into \{lang\} syntax.\\
2. Model the intention written in natural language using \{lang\}.\\
3. Express the requirement using \{lang\}.\\
4. Model the given natural language into \{lang\}.\\
5. Translate the given requirement using \{lang\}'s syntax and semantics.\\

(Randomly choose one of the above.)\\

You only need to return the \{lang\} formal specification without explanation.\\

\# \textbf{Input}\\
<Input goes here>

\end{tcolorbox}
\caption{Prompt for SegGen Task.}
\label{prompt:seggen}
\end{figure*}

\begin{figure*}[th]
\begin{tcolorbox}[colframe=cyan!40!black, title=\textbf{Prompt for ProofGen Task (lang: a placeholder to be replaced by each formal specification language name.)}]

\# \textbf{Task Description (ProofGen)}

1. Translate the given requirements into {lang} syntax.\\
2. Model the given requirements written in natural language using {lang}.\\
3. Express the requirements using {lang}.\\
4. Model the given requirements written in natural language into {lang}.\\
5. Translate the given requirements into {lang}'s syntax and semantics.\\

(Randomly choose one of the above.)\\

(For ACSL): You only need to return the {lang} formal specification with the code without explanation.

(For others): You only need to return the {lang} formal specification without explanation.\\

\# \textbf{Input}\\
<Input goes here>

\end{tcolorbox}
\caption{Prompt for ProofGen Task.}
\label{prompt:proofgen}
\end{figure*}

\begin{figure*}[th]
\begin{tcolorbox}[colframe=cyan!40!black, title=\textbf{Prompt for ProofComplete Task (lang: a placeholder to be replaced by each formal specification language name.)}]

\# \textbf{Task Description (ProofComplete)}

1. Please complete the following formal proof in formal specification language {lang} according to the given requirement.\\
2. Please complete the following formal proof in {lang} according to the given requirement.\\
3. Please complete the given formal proof in {lang} following the requirement below.\\
4. Please complete the following formal proof in {lang} according to the requirement below.\\
5. Please complete the following formal proof in {lang} according to the given requirement.\\

(Randomly choose one of the above.)\\

You only need to return the completed {lang} formal specification (together with the provided formal specification) without explanation.\\

\# \textbf{Input}\\
<Input goes here>

\end{tcolorbox}
\caption{Prompt for ProofComplete Task.}
\label{prompt:proofcomplete}
\end{figure*}

\begin{figure*}[th]
\begin{tcolorbox}[colframe=cyan!40!black, title=\textbf{Prompt for ProofInfill Task (lang: a placeholder to be replaced by each formal specification language name.)}]

\# \textbf{Task Description (ProofInfill)}

1. Please fill in the [MASK] in the following formal proof in formal specification language {lang} according to the given requirement.\\
2. Please fill in the [MASK] in the following formal proof in {lang} according to the given requirement.\\
3. Please complete the given formal proof in {lang} following the requirement below by filling in the [MASK].\\
4. Please fill in the [MASK] in the following formal proof in {lang} according to the requirements below.\\
5. Please fill in the [MASK] in the following formal proof in {lang} according to the given requirement.\\

(Randomly choose one of the above.)\\

You only need to return the completed {lang} formal specification (together with the provided formal specification) without explanation.\\

\# \textbf{Input}\\
<Input goes here>

\end{tcolorbox}
\caption{Prompt for ProofInfill Task.}
\label{prompt:proofinfill}
\end{figure*}

\begin{figure*}[th]
\begin{tcolorbox}[colframe=cyan!40!black, title=\textbf{Prompt for Code2Proof Task}]

\# \textbf{Task Description (Code2Proof)}

1. Please fill in the [MASK] in ACSL according to the given requirement and ACSL specification.\\
2. Please fill in the [MASK] in ACSL according to the given requirement.\\
3. Please fill in the [MASK] in ACSL according to the given ACSL specification.\\
4. Please fill in the [MASK] in ACSL according to the given requirement and ACSL specification.\\
5. Please infill the [MASK] in ACSL according to the given requirement.\\

(Randomly choose one of the above.)\\

You only need to return the completed ACSL formal specification (together with the provided formal specifications and C programs) without explanation.\\

\# \textbf{Input}\\
<Input goes here>

\end{tcolorbox}
\caption{Prompt for Code2Proof Task.}
\label{prompt:code2proof}
\end{figure*}